\documentclass[]{techreport}

\renewcommand{\reportshorttitle}{World-State Reasoning in Video Continuation}
\renewcommand{\reportdate}{September 2026}
\newcommand{\reportpdftitle}{Do Video Generators Track the World Across Segments? A Benchmark and Method for World-State Reasoning in Video Continuation}
\newcommand{\reportpdfauthor}{Yingmao Miao, Pengfei Zhang, Chaoran Xu, Meng Yu, Jing Tang, Xiangxiang Chu, Chao Shen, and Chenhao Lin}

\setreportlogo{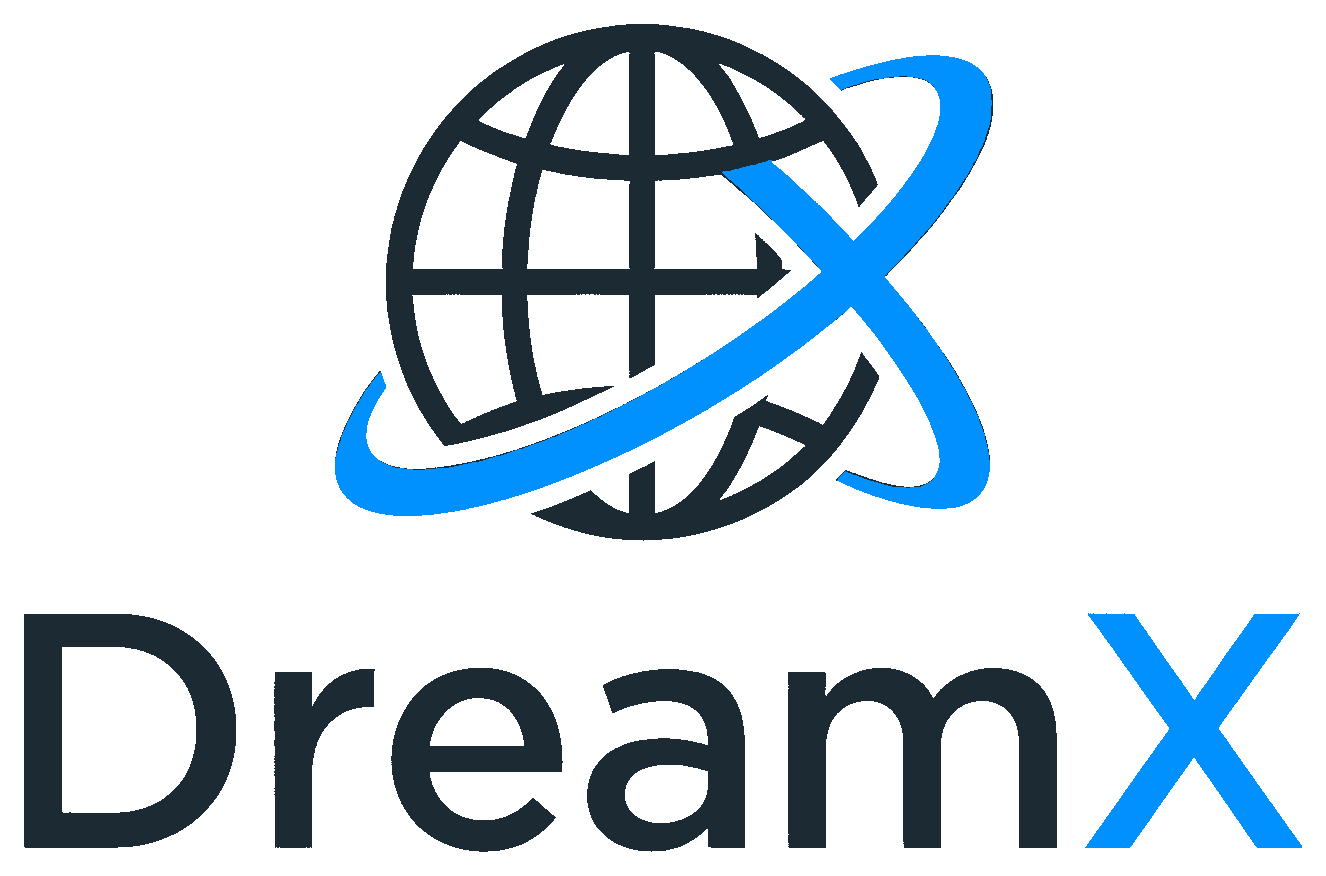}

\title{Do Video Generators Track the World Across Segments? A Benchmark and Method for World-State Reasoning in Video Continuation}

\author[1,2,\dagger]{Yingmao Miao}
\author[2,\ddagger]{Pengfei Zhang}
\author[2,\dagger]{Chaoran Xu}
\author[2,3,\dagger]{Meng Yu}
\author[2,*]{Jing Tang}
\author[2]{Xiangxiang Chu}
\author[1]{Chao Shen}
\author[1,*]{Chenhao Lin}
\affiliation[1]{Xi'an Jiaotong University}
\affiliation[2]{DreamX Team, Alibaba Group}
\affiliation[3]{Shanghai Jiaotong University}
\contribution[\dagger]{Work done during an internship at DreamX Team, Alibaba Group}
\contribution[\ddagger]{Project leader}
\contribution[*]{Corresponding authors.}

\abstract{Video generators build long videos by composing shorter parts, either by generating segments one after another or by autoregressively extending chunks. Each new part usually depends on memories of historical observations, such as recent frames, selected key frames, memory banks, or cached features.
These memories preserve visible evidence from the past, but current generators do not reliably turn such evidence into a world-state interface: what holds in the video world after previous actions and how it should change under the next prompt. A past frame remains valid history, but it may not describe the state needed by the next segment; some states must instead be inferred from occluded or implicit changes rather than copied from a directly observed frame.
This creates a simple but overlooked question for video continuation: given a previous video, its prompt, and a new prompt, can a model generate a continuation that reflects the state determined by both the historical video and the new prompt? To answer this question, we introduce \statebench, a benchmark that targets this gap by testing continuations over three state categories: past-visible states, occluded-process states, and complex-transition states. We further propose \stateagent, which explicitly maintains an entity-state representation, updates it under the new prompt, grounds the predicted post-action state as a future end frame, and renders the next video.
Experiments show that our method improves controlled video continuation by raising the all-case state score (SCS-All) from 45.2 to 69.3, and also benefits story generation at the one-minute scale. Code is avaliable at \url{https://github.com/AMAP-ML/StateAgent}.
}

\date{\reportdate}

\usepackage{amsmath}
\usepackage{amssymb}
\usepackage{array}
\usepackage{bm}
\usepackage{colortbl}
\usepackage{enumitem}
\usepackage{xspace}
\usepackage{fancyhdr}
\usepackage{float}
\usepackage{wrapfig}
\usepackage{xurl}
\definecolor{BrandPurple}{HTML}{7366CC}
\definecolor{BrandCyan}{HTML}{00B4E5}
\definecolor{LightGray}{HTML}{F5F5F5}
\definecolor{TableRowAlt}{HTML}{EAF4FB}
\definecolor{HeaderGray}{HTML}{888888}
\definecolor{ForestGreen}{HTML}{228B22}
\definecolor{Gray}{HTML}{808080}

\newcolumntype{L}[1]{>{\raggedright\arraybackslash}m{#1}}
\newcolumntype{C}[1]{>{\centering\arraybackslash}m{#1}}

\newcommand{\statebench}{\textsc{StateBench}\xspace}
\newcommand{\stateagent}{\textsc{StateAgent}\xspace}

\newtcolorbox{highlight}{
    colback=BrandCyan!5,
    colframe=BrandCyan,
    arc=2pt,
    boxrule=0.8pt,
    left=6pt, right=6pt, top=4pt, bottom=4pt,
    breakable
}

\newtcolorbox{keyfinding}{
    colback=BrandPurple!5,
    colframe=BrandPurple,
    coltitle=white,
    fonttitle=\bfseries,
    title=Key Finding,
    arc=2pt,
    boxrule=1pt,
    left=6pt, right=6pt, top=4pt, bottom=4pt,
    breakable
}

\newtcolorbox{tipbox}[1]{
    colback=ReportSurface,
    colframe=ReportPrimary,
    coltitle=white,
    fonttitle=\bfseries,
    title=#1,
    arc=2pt,
    boxrule=1pt,
    left=6pt, right=6pt, top=4pt, bottom=4pt,
    breakable
}

\newtcolorbox{promptbox}[1][]{
    enhanced,
    breakable,
    colback=gray!10,
    colframe=black,
    arc=3pt,
    boxrule=0.5pt,
    left=8pt, right=8pt, top=8pt, bottom=8pt,
    fontupper=\small,
    title={#1},
    coltitle=black,
    colbacktitle=gray!10,
    attach boxed title to top left={yshift=-2mm, xshift=5mm},
    boxed title style={boxrule=0.5pt, colframe=black, arc=2pt}
}

\fancypagestyle{plain}{%
    \fancyhf{}
    \fancyhead[L]{\footnotesize\color{HeaderGray}\sffamily \reportshorttitle}
    \fancyhead[R]{\footnotesize\color{HeaderGray} \thepage}
    \fancyfoot[C]{}

}

\fancypagestyle{firstpage}{%
    \fancyhf{}
    \fancyhead[L]{\small\sffamily\color{ReportPrimary}
      DreamX Team}
    \fancyhead[R]{\small\color{HeaderGray} \reportdate}
    \fancyfoot[C]{\footnotesize\color{gray}\thepage}

}

\hypersetup{
  pdftitle={\reportpdftitle},
  pdfauthor={\reportpdfauthor}
}

\begin{document}

\maketitle
\thispagestyle{firstpage}

\section{Introduction}

As video generation aims for longer outputs and coherent visual stories, current systems face two practical choices. One is to generate a very long video in a generation process, which is computationally expensive and remains difficult for current models. The other is to generate shorter segments or chunks sequentially, where each new unit must be conditioned on the generated history. Providing the complete history to every new unit is also costly and often impractical, so existing continuation systems usually compress or select the history through recent frames, selected key frames~\citep{storydiffusion,storymem}, visual memory banks~\citep{videomemory,worldmem,vmem,spatialmemory}, or cached features~\citep{contextforcing,causalcine,DreamX-world}. These signals form an observation memory of the generated history, useful for preserving appearance, identity, etc.

\begin{figure}[t]
\centering
\includegraphics[width=0.95\linewidth]{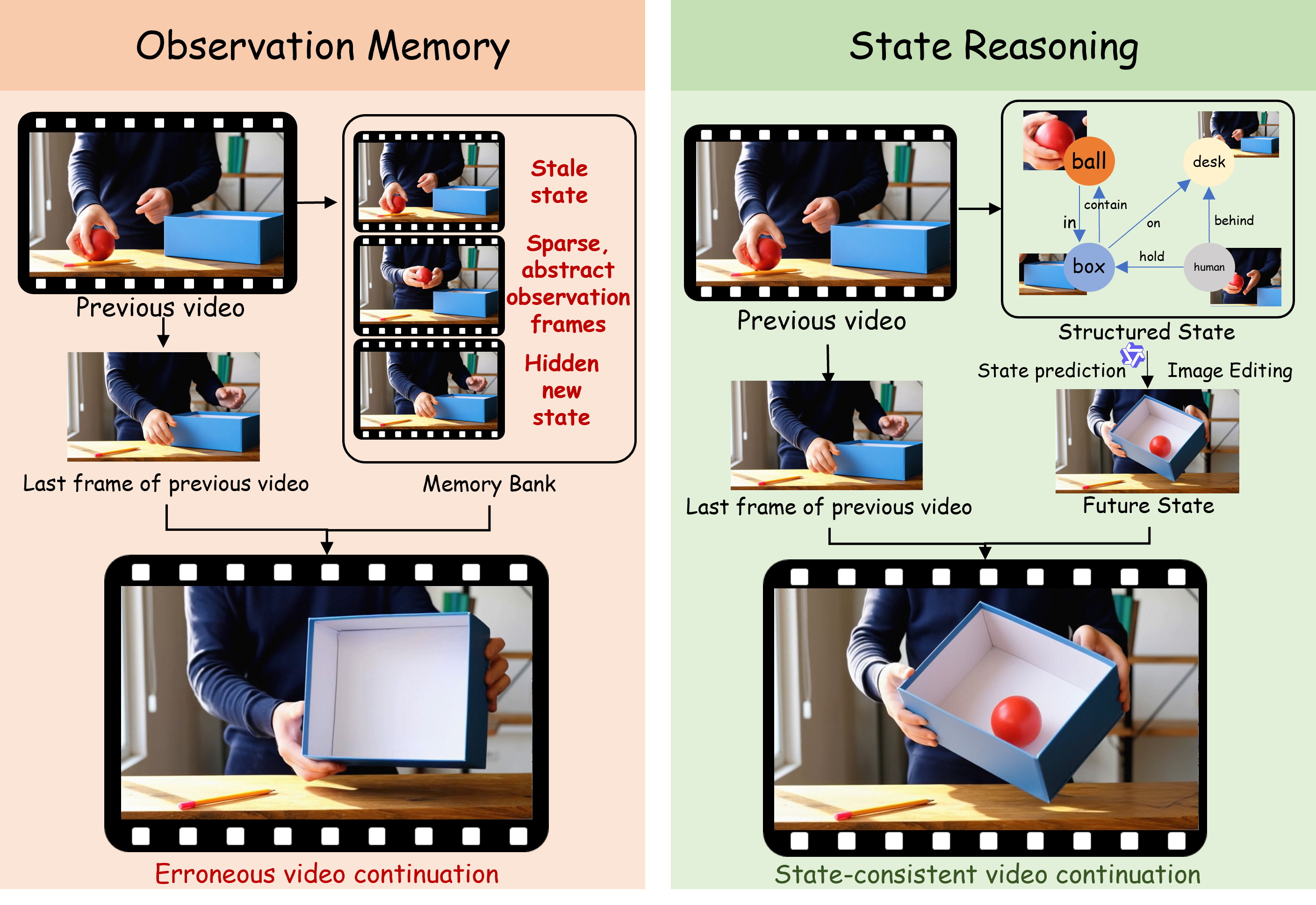}
\caption{\textbf{Cross-segment state reasoning.} Observation memory stores visible evidence from previous frames, while state reasoning updates what should now be true after the historical action. For example, after an object is placed into a container and becomes invisible, the next segment should know the object is still inside, even if the boundary frame only shows the container.}
\label{fig:overview}
\end{figure}

Observation memory may contain evidence for the world state, but current continuation methods expose it as frames or features rather than as an updated state interface. Historical frames record what was visible at particular moments, while the next segment needs what should hold after the previous video and how it should change under the new prompt. A past frame is valid history, but not always current-state evidence: the relevant state may be hidden at the boundary, changed after an earlier observation, or inferred from an occluded process. We view this evidence-to-state conversion as a capability future video world models should internalize; under current model capabilities, we make it explicit through state reasoning.

Several recent works have started to study hidden state, out-of-sight content, or long-horizon world dynamics, but they address different settings from video continuation. StEvo-Bench studies hidden state evolution within a single generated video, but it requires the model to establish, hide, evolve, and reveal the state in one generation, making failures hard to attribute~\citep{stevobench}. PAN uses LLM latent features as world states for long-horizon dynamics. This is useful for abstract simulation, but its state lacks explicit appearance binding, making it difficult to preserve the visual identity of entities in the next generated segment~\citep{pan}. LiveWorld stores persistent 4D spatiotemporal memory for out-of-sight content, which can preserve rich scene evidence but is costly and difficult to scale. Crucially for our setting, it still stores past observations in a richer space, rather than explicitly converting historical events into an updated state for the next video segment~\citep{liveworld}. These differences leave open how to evaluate and support state update in video continuation.

We study the basic unit of state-aware video continuation. Given a previous video $V_t$, its source prompt $p_t$, and a continuation prompt $p_{t+1}$, a model should infer the world state at time $t+1$ by updating the historical state according to the continuation prompt, rather than merely maintaining visual continuity with previous frames. We introduce \statebench to operationalize this setting. \statebench contains 200 tasks across three state categories: past-visible states, occluded-process states, and complex-transition states. For each task, we provide a task-specific checklist. Evaluation first checks whether the generated continuation completes the state-establishing action requested by the prompt, and then checks whether the resulting state is correct.

We further propose \stateagent as a practical solution under current model capabilities. Our key insight is that state-consistent continuation should decouple state reasoning from video rendering: an explicit state-reasoning role should decide the next state, while the video generator should render that predicted state into pixels. \stateagent implements this idea by building an entity-state representation from the previous video, updating it under the continuation prompt to obtain the post-action state, grounding that state as a future end frame using an image editor, and rendering the continuation. Experiments show that stronger observation memory does not make current generators reliably capture the required world state. Recent-frame, key-frame, memory-bank, and autoregressive baselines often preserve visible appearance, but still fail when the current state is hidden, changed after an earlier observation or produced by an unseen change. Overall, our contributions are:

\begin{itemize}[leftmargin=*,topsep=2pt,itemsep=1pt]
    \item We identify the gap between observation memory and state reasoning in video continuation and introduce \statebench, a unit benchmark that tests whether generated continuations follow the state implied by the previous video and the new prompt.
    \item We propose \stateagent, which decouples state updating from video rendering by leveraging a VLM to parse historical states and predict future states, grounding them into future frames via an image editor, and rendering the video continuation through keyframe-to-video generation.
    \item We show that our method outperforms observation-based memory methods, not only improving controlled video continuation by raising the \statebench score from 45.2 to 69.3, but also benefiting long story generation.
\end{itemize}

\section{Problem Formulation}

\subsection{Cross-Segment State Handoff}

\begin{figure}[t]
\centering
\includegraphics[width=\linewidth]{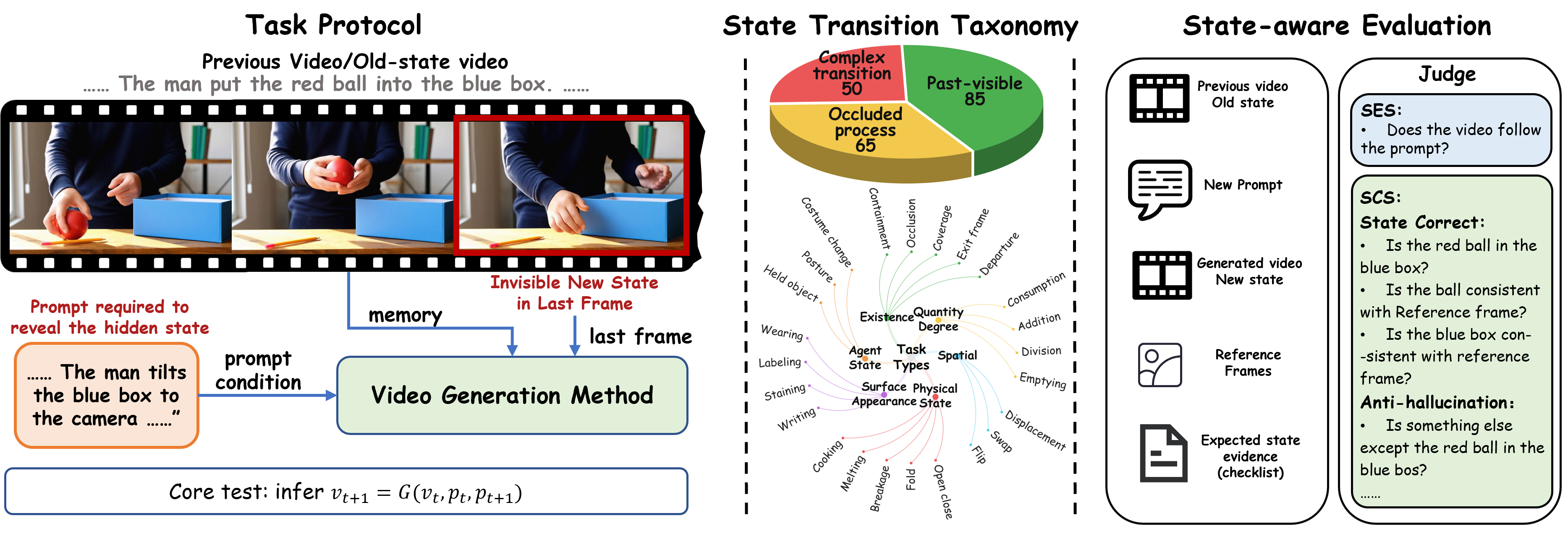}
\caption{\textbf{\statebench overview.} Given a previous video $V_t$, its source prompt $p_t$, and a continuation prompt $p_{t+1}$, a method must generate $V_{t+1}$ that reveals or uses the state established in the history. \textbf{Left: task protocol.} The previous segment contains the state-establishing event, while its tail frame, i.e., the generation boundary between two segments, hides or ambiguates the task-relevant relation. \textbf{Middle: state-evidence categories.} \statebench evaluates continuations across different ways in which the target state is evidenced by the history. \textbf{Right: State-aware evaluation.} SES first checks whether the requested reveal/revisit action is completed, and SCS then measures whether the exposed state is consistent with the previous video.}
\label{fig:statebench}
\end{figure}

Let $V_t$ be a previous video, $p_t$ its source prompt, and $p_{t+1}$ the prompt for the next segment. Current continuation interfaces usually pass history forward through an observation-memory interface $O_t=M_{\mathrm{obs}}(V_t)$, such as the last frame, recent frames, retrieved key frames, memory-bank entries, or cached temporal features. Here $M_{\mathrm{obs}}$ denotes the history interface actually exposed to the next generator, rather than the full information content of $V_t$. This interface exposes visual evidence in sparse, compressed, or potentially stale forms, making it difficult for current generators to infer a state that can be queried, updated, and grounded for rendering. The next segment is then generated as:
\begin{equation}
    V_{t+1} \sim G(O_t,\; p_{t+1}).
\end{equation}

The problem is therefore an interface gap. At the segment boundary, the observation exposed to the next generator may not make the post-history state explicit. In our setting, this boundary is the tail frame of the previous video: the historical event has established a task-relevant state, but the tail frame hides or ambiguates the relation. A state-aware continuation process should therefore preserve both visual evidence and an explicit post-history state belief:
\begin{equation}
    S_t = E(V_t,p_t), \qquad
    V_{t+1} \sim G(O_t,\; S_t,\; p_{t+1}),
\end{equation}
where $O_t$ preserves visual evidence such as appearance, identity, layout, and style, while $S_t$ denotes what should be physically true after the previous video. The central cross-segment state handoff problem occurs when the observation memory interface maps different histories (a) and (b) to similar exposed evidence, while their post-history states differ:
\begin{equation}
M_{\mathrm{obs}}(V_t^{(a)}) \approx M_{\mathrm{obs}}(V_t^{(b)}), \qquad S_t^{(a)} \neq S_t^{(b)}.
\end{equation}
A generator that relies only on the exposed observation memory interface can struggle to distinguish these cases under current model capabilities, while an explicit state representation can condition the next segment on what is currently true.

\subsection{State as Causal Belief}

We define a world state as a set of task-relevant physical relations that may persist beyond visibility. The previous video and its prompt establish a post-history state $S_t$. The continuation prompt then induces a state update that may reveal, use, or further transform this state. We write this state reasoning step as:
\begin{equation}
    S_{t+1} = \tau(S_t,\; a_{t+1}),
\end{equation}
where $a_{t+1}$ is the physical event described or implied by $p_{t+1}$. Here, $S_{t+1}$ denotes the state that should be reflected in the continuation, and $\tau$ abstracts how the continuation prompt uses or updates the state carried from the previous segment. Cross-segment generation therefore requires inferring $S_t$ from the historical event and generating $V_{t+1}$ according to the state implied by both $S_t$ and $p_{t+1}$.

In \statebench, we instantiate $S_t$ as a set of task-relevant physical relations that expose the gap between observation and state:
\begin{equation}
S_t = \left\{ r_i \right\}, \qquad r_i \in \mathcal{F},
\end{equation}
where $\mathcal{F}$ denotes the state types evaluated in \statebench, including visibility, location, containment, transformation, surface and material state, quantity, and agent-object relations. These variables provide controlled tests of whether a continuation preserves the physical state implied by the previous segment.

\subsection{Observation Memory Versus State Reasoning}

Observation memory asks whether generated content can reuse or resemble previous visible evidence; state reasoning asks whether current generators can infer and maintain the physical consequences of previous actions for the next segment. As illustrated in Figure~\ref{fig:overview}, appearance continuity and state continuity are therefore complementary but distinct. A visual memory mechanism can preserve how entities looked, where they appeared, and how the scene was framed, but it is not a reliable state interface for deciding what should now be true after an action. This distinction motivates \statebench, where continuations are evaluated not only by whether they follow the next prompt, but also by whether the revealed state is consistent with the previous video.

\section{\statebench: Measuring State Discontinuity}

\statebench is a compact unit test for state-aware video continuation. Each task provides a 5--10s previous video $V_t$, its source prompt $p_t$, and a continuation prompt $p_{t+1}$. The previous segment establishes a physical state, but its tail frame hides or ambiguates the task-relevant relation. The continuation must therefore use the state established before the segment boundary, rather than only the terminal appearance, to complete the requested state update.

\subsection{Task and Dataset}

\statebench contains 200 cross-segment continuation cases across three state-evidence categories: 85 past-visible cases, 65 occluded-process cases, and 50 complex-transition cases. In every case, the task-relevant state is hidden or ambiguous at the generation boundary, and the next prompt induces an action that exposes, uses, or transforms this state without naming the target answer. Past-visible cases contain a frame that directly shows the target state before it becomes hidden. Occluded-process cases involve state changes inside a container, behind an obstruction, or otherwise out of direct view. Complex-transition cases involve harder physical or relational state transitions, such as spatial transfer, quantity change, color mixing, dissolution, or material and surface transformations.
Figure~\ref{fig:statebench} summarizes the task protocol, category split, and evaluation design. Detailed schemas, fine-grained categories, and prompt construction rules are provided in the appendix.

\subsection{Evaluation}

Each generation method receives the same triplet $(V_t,p_t,p_{t+1})$ and generates one continuation $\hat{V}_{t+1}$. The evaluator receives $(V_t,p_t,p_{t+1},\hat{V}_{t+1})$, a task-specific checklist, and reference frames selected from $V_t$ for appearance-sensitive entities. These reference frames are used only for evaluation, not as additional generation inputs.

We use a two-stage checklist evaluation. \emph{State Exposure Score} (SES) first measures whether the continuation completes the requested state-checking or state-updating action. For SES-positive samples, the state checklist contains state-correctness questions and anti-hallucination questions. Let $e_i\in\{0,1\}$ indicate successful exposure, and let $y_{iq}\in\{0,1\}$ denote whether the judge answers ``yes'' to checklist question $q$ for sample $i$:
\begin{equation}
\mathrm{SES}=\frac{1}{N}\sum_{i=1}^{N}e_i.
\end{equation}

For SES-positive samples, \emph{State Correctness} (SC) is the yes rate over state-correctness questions, while \emph{Hallucination Rate} (HR) is one minus the yes rate over anti-hallucination questions:
\begin{equation}
\mathrm{SC}=\frac{\sum_i e_i\sum_{q\in\mathcal{C}_i} y_{iq}}{\sum_i e_i|\mathcal{C}_i|},
\qquad
\mathrm{HR}=1-\frac{\sum_i e_i\sum_{q\in\mathcal{H}_i} y_{iq}}{\sum_i e_i|\mathcal{H}_i|}.
\end{equation}
The conditional state consistency score is the yes rate over all state-checklist questions on SES-positive samples:
\begin{equation}
\mathrm{SCS}_{\mathrm{cond}}=
\frac{\sum_i e_i\sum_{q\in\mathcal{C}_i\cup\mathcal{H}_i} y_{iq}}{\sum_i e_i|\mathcal{C}_i\cup\mathcal{H}_i|}.
\end{equation}
We also report an all-case score, SCS-All, which treats exposure failure as an end-to-end state failure:
\begin{equation}
\mathrm{SCS}_{\mathrm{All}}=\frac{1}{N}\sum_i e_i c_i,
\qquad
c_i=\frac{1}{|\mathcal{C}_i\cup\mathcal{H}_i|}\sum_{q\in\mathcal{C}_i\cup\mathcal{H}_i} y_{iq}.
\end{equation}
In the tables, SCS denotes $\mathrm{SCS}_{\mathrm{cond}}$ for readability.

Following criterion-specific VLM evaluation protocols used in other benchmarks~\citep{videophy,phygenbench,stevobench}, scores are produced by a checklist judge. The judge is given the previous video, prompts, generated continuation, task checklist, and entity reference frames when appearance matching is required.

\section{\stateagent: Future-Frame State Carrier}

\begin{figure}[t]
\centering
\includegraphics[width=\linewidth]{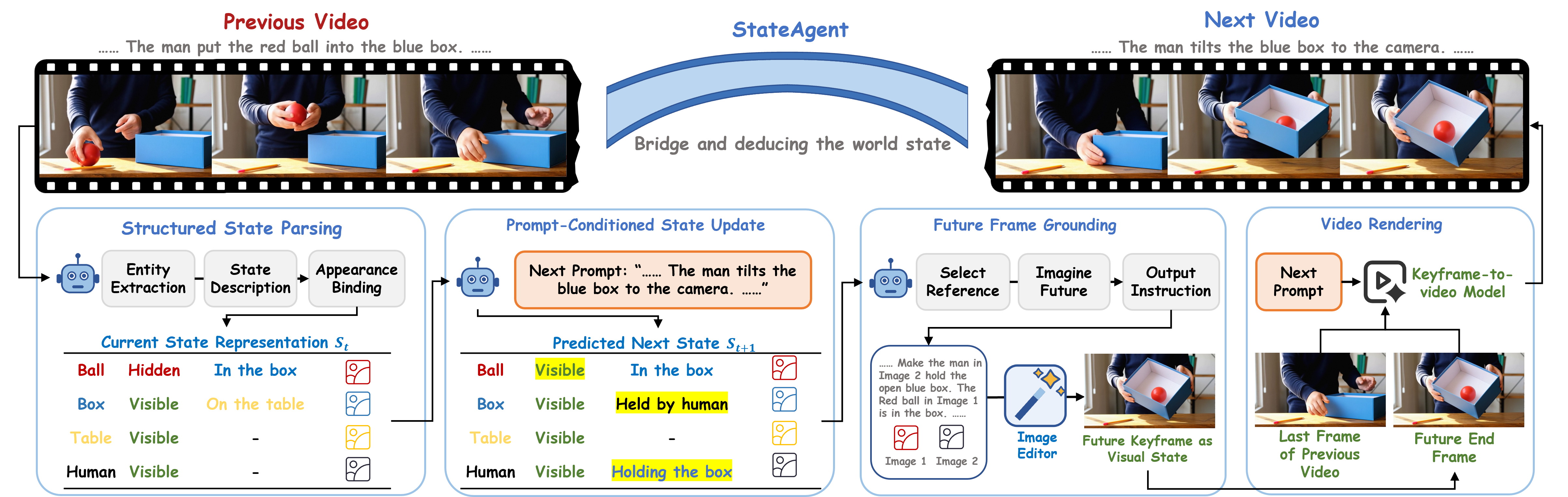}
\caption{\textbf{\stateagent pipeline.} \stateagent decouples state prediction from video rendering. Given $V_t$, $p_t$, and $p_{t+1}$, it parses the historical event into a structured state graph, predicts the target post-action state, grounds it into a future end frame through image editing, and renders the continuation with first--last-frame video generation.}
\label{fig:pipeline}
\end{figure}

\statebench defines a capability that future video generation models should eventually internalize: carrying a world state across generation boundaries and updating it under a new prompt. Current video generators, however, do not reliably expose such a state interface. We therefore present \stateagent as a practical solution for the current stage of model capability. The method decouples \emph{state prediction} from \emph{state rendering}: an explicit predictor first infers what should be true after the continuation prompt, and a video generator then renders a segment constrained by that predicted state.
\stateagent instantiates this state handoff formulation by making the intermediate state update explicit as shown in Figure~\ref{fig:pipeline}. The parsed state graph represents $S_t$, the state after the previous segment. Given the continuation prompt, \stateagent predicts a target continuation state $\hat{S}_{t+1}$, grounds it as a future end frame, and then renders the next segment toward that predicted state.

This decoupling raises a representation question. Purely textual states are easy to update but too abstract to preserve appearance bindings; implicit latent states may support dynamics, but are difficult to inspect, costly to train, and can discard visual details needed for faithful continuation. \stateagent instead uses a \emph{future end frame} as a visualized state carrier. A multimodal agent predicts the structured post-action state, an image-editing model grounds that state into pixels with historical appearance references, and a keyframe-to-video generator renders the motion between the historical end frame and the predicted future end frame. This follows a similar intuition to ImageWAM, which uses image editing as a compact target-state transformation prior for world-action modeling~\citep{zhang2026imagewam}.

\subsection{Structured State Carrier}

Given $(V_t,p_t)$, \stateagent builds a lightweight state graph $S_t$ rather than retrieved frames. Each entity node stores an identity, references to frames where the entity is visible, and a short state description that summarizes its current condition, such as whether it is open, broken, hidden, held, or stained. Relations record visibility, containment, location and so on. The graph is deliberately sparse: it stores only the state variables needed for future continuation, while leaving texture, pose, and motion synthesis to the renderer. Unlike key-frame memory, it can mark a relation as true even when not visible at the boundary, and it can mark an earlier visual observation as stale after a later action changes the state. When constructing $S_t$, \stateagent uses $p_t$ only for entity binding between names and visual entities, while state parsing is grounded in the observed video evidence.

\subsection{Prompt-Conditioned State Update}

The continuation prompt is treated as an action condition that induces a state update. Given the parsed post-history state $S_t$ and $p_{t+1}$, a state updater $\mathcal{U}$ predicts the target continuation state:
\begin{equation}
\hat{S}_{t+1} = \mathcal{U}(S_t,\; p_{t+1}).
\end{equation}
Here, $\hat{S}_{t+1}$ is the method's explicit estimate of the state that should be reflected in $V_{t+1}$. It may reveal a hidden relation, use an existing relation, or further transform the physical state. The output specifies the target state for the continuation.
This directly addresses the interface bottleneck of observation memory: hidden objects can remain in $S_t$, while stale appearances from earlier frames are not treated as the current state after later actions.

\subsection{Future-Frame Grounding}

A structured state alone is not enough to condition a video generator, because it lacks concrete appearance binding. \stateagent therefore grounds $\hat{S}_{t+1}$ into a future end frame $\hat{F}_{t+1}^{\mathrm{end}}$. This step converts the symbolic state prediction into a visual condition that current video generators can follow. The historical end frame $F_t^{\mathrm{end}}$ provides camera, layout, and visible entities; reference frames $R_t$ provide appearances for entities that are not visible at the boundary but are needed in the predicted state. A frame-grounding module $\mathcal{G}_{\mathrm{frame}}$ receives the historical end frame, selected references, the predicted state, and the continuation prompt:
\begin{equation}
\hat{F}_{t+1}^{\mathrm{end}} =
\mathcal{G}_{\mathrm{frame}}(F_t^{\mathrm{end}}, R_t, \hat{S}_{t+1}, p_{t+1}).
\end{equation}
The editing instruction is generated from the predicted state change and the selected visual references, so that the edited frame reflects the target state while preserving appearance and layout. The target frame is thus not a memory frame; it is a visualized prediction of the future state.

\subsection{Video Rendering and Diagnostics}

Finally, a keyframe-to-video generator $\mathcal{G}_{\mathrm{video}}$ renders the continuation:
\begin{equation}
\hat{V}_{t+1}=
\mathcal{G}_{\mathrm{video}}(F_t^{\mathrm{end}},\hat{F}_{t+1}^{\mathrm{end}},p_{t+1}),
\end{equation}
where we instantiate $\mathcal{G}_{\mathrm{video}}$ with the keyframe-to-video interface of Wan~\citep{wan}. The historical end frame keeps temporal continuity with the previous segment, while the future end frame constrains the state that must be reflected by the continuation. The renderer is therefore responsible for motion and visual realism, rather than rediscovering hidden physical state from scratch.

The modular structure makes failures inspectable. A wrong continuation can be attributed to history parsing, state update, future-frame grounding, or video rendering. This is useful because black-box video generation failures often look identical in the final video, while \stateagent exposes whether the system failed to carry the state, update it, bind it to appearance, or animate toward it.

\section{Experiments}

\subsection{Experimental Setup}

\stateagent uses Qwen3.5-Plus~\citep{yang2025qwen3technicalreport} as the parser for historical states, the state updater, and the editing-instruction generator, Wan2.7-Image~\citep{mao2026wanimagepushingboundariesgenerative} as the future-frame generator and Wan2.2-KF2V-Flash as the state-to-video renderer. We evaluate all methods under the same input $(V_t,p_t,p_{t+1})$. Each method generates only the next segment; the previous video is used only as conditioning evidence. Unless otherwise specified, output length, resolution, prompt template, sampling settings, and random seeds are matched within each model family. We report SES and SCS across the three \statebench categories. For diagnosis, we also report State Consistency (SC) and Hallucination Rate (HR) within each category; the Overall columns report aggregate SES, conditional SCS, and SCS-All.

\paragraph{Main baselines.}
The main comparison covers three families of continuation interfaces. \emph{Frame-conditioned baselines} include I2V from the final frame and a recent-frame~\citep{longcat} variant that conditions on the last few frames. \emph{Key-frame and memory baselines} include StoryMem~\citep{storymem} and StoryMem supplied with full key frames covering the complete state-establishing process. \emph{Autoregressive baselines} include MAGI~\citep{magi1}.

\subsection{Main Results on \statebench}

\begin{figure}[t]
\centering
\includegraphics[width=0.94\linewidth]{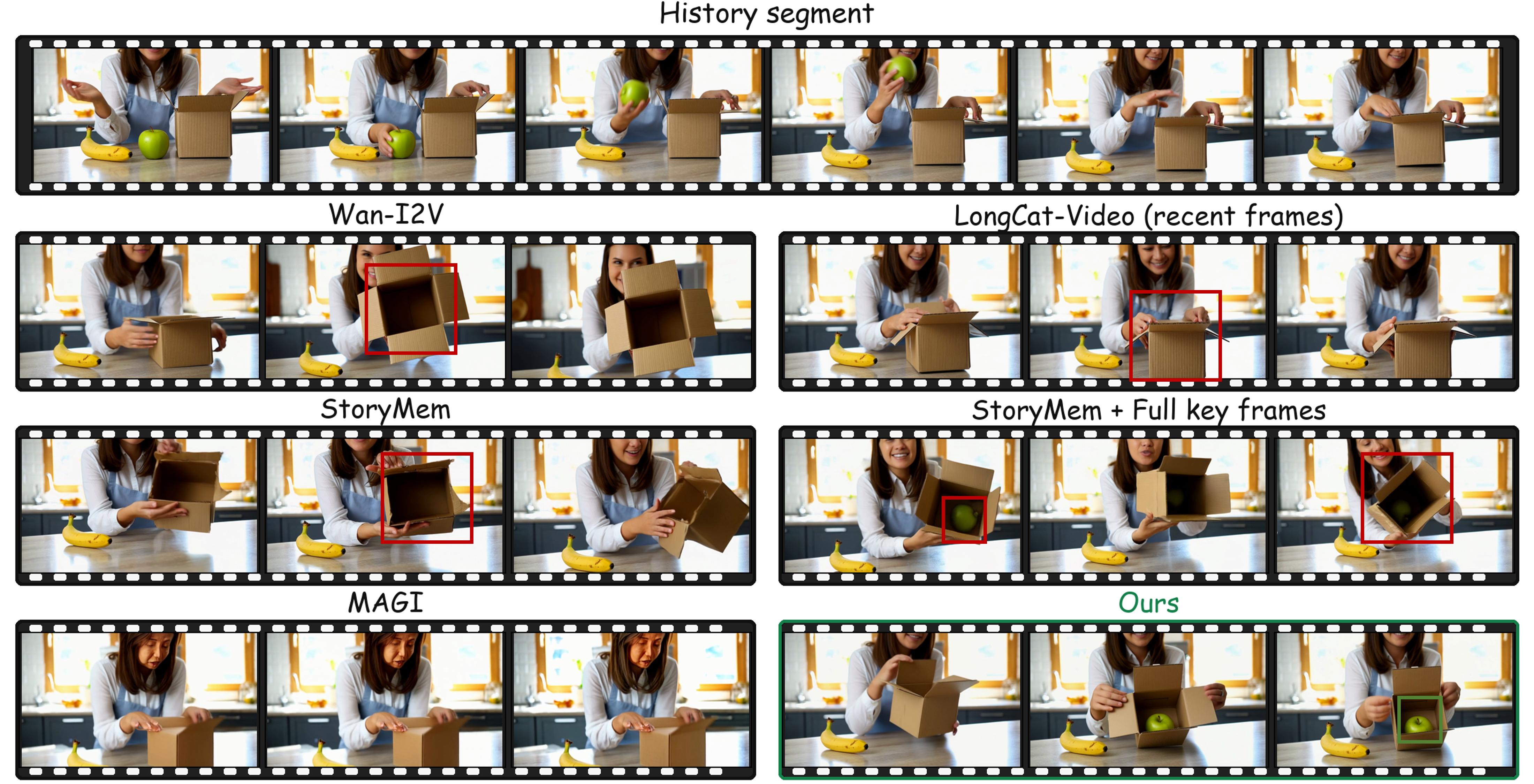}
\caption{\textbf{Example on \statebench.} Setup prompt: a woman drops a green apple into an open cardboard box while a banana remains on the kitchen counter. Continuation prompt: ``A single continuous realistic video on a kitchen counter. The woman tilts the cardboard box toward the camera to show what is inside. Fixed camera, medium shot, bright indoor lighting.''}
\label{fig:qualitative}
\end{figure}

\begin{table}[t]
\centering
\small
\setlength{\tabcolsep}{3pt}
\renewcommand{\arraystretch}{0.95}
\resizebox{\linewidth}{!}{%
\begin{tabular}{@{}lccccccccccccccc@{}}
\toprule
& \multicolumn{4}{c}{Past-visible} & \multicolumn{4}{c}{Occluded-process} & \multicolumn{4}{c}{Complex-transition} & \multicolumn{3}{c}{Overall} \\
\cmidrule(lr){2-5}\cmidrule(lr){6-9}\cmidrule(lr){10-13}\cmidrule(lr){14-16}
Method & SES$\uparrow$ & SC$\uparrow$ & HR$\downarrow$ & SCS$\uparrow$ & SES$\uparrow$ & SC$\uparrow$ & HR$\downarrow$ & SCS$\uparrow$ & SES$\uparrow$ & SC$\uparrow$ & HR$\downarrow$ & SCS$\uparrow$ & SES$\uparrow$ & SCS$\uparrow$ & SCS-All$\uparrow$ \\
\midrule
Wan-I2V & 69.4 & 26.0 & 40.7 & 42.9 & \underline{90.8} & 22.6 & 49.2 & 36.1 & \underline{76.0} & 17.1 & 18.4 & 48.2 & 78.0 & 41.6 & 32.5 \\
LongCat-Video (recent frames) & 57.6 & 27.9 & 22.4 & 51.8 & 70.8 & 23.9 & 27.2 & 46.0 & 60.0 & 10.6 & \textbf{10.0} & 49.2 & 62.5 & 49.0 & 30.6 \\
StoryMem~\citep{storymem} & 80.0 & 50.5 & 15.4 & 66.9 & 75.4 & 34.7 & 20.4 & 54.8 & 66.0 & 27.8 & 34.8 & 46.5 & 75.0 & \underline{58.5} & 43.9 \\
StoryMem + full KFs & \textbf{91.2} & \underline{51.9} & \underline{13.5} & \underline{68.3} & 86.0 & \underline{36.1} & \underline{19.4} & \underline{55.5} & 56.0 & 19.0 & 37.5 & 41.1 & \underline{78.5} & 57.5 & \underline{45.2} \\
MAGI-1~\citep{magi1} & 40.0 & 19.6 & 23.5 & 47.1 & 40.0 & 21.8 & 51.9 & 33.5 & 12.0 & \underline{44.4} & 25.0 & \underline{59.2} & 33.0 & 42.8 & 14.1 \\
\stateagent & \underline{89.4} & \textbf{61.0} & \textbf{11.2} & \textbf{74.1} & \textbf{96.9} & \textbf{68.8} & \textbf{10.3} & \textbf{78.5} & \textbf{92.0} & \textbf{61.2} & \underline{17.4} & \textbf{71.4} & \textbf{92.5} & \textbf{74.9} & \textbf{69.3} \\
\bottomrule
\end{tabular}%
}
\caption{\textbf{Main \statebench results.} SES measures completion of the requested state-exposing or state-updating action; SCS denotes conditional state consistency on SES-positive samples. SC is the yes rate over state-correctness questions, while HR is the no rate over anti-hallucination questions. SCS-All is the all-case state score and treats exposure failure as failure.}
\label{tab:main_results}
\end{table}

Table~\ref{tab:main_results} evaluates whether stronger visual history interfaces let current generators recover the state needed for cross-segment handoff. Although observation-memory methods often complete the action described by the prompt, they frequently expose a state inconsistent with the previous segment. \stateagent improves conditional SCS from 58.5 to 74.9 over the strongest baseline, and improves the all-case SCS-All from 45.2 to 69.3, showing that the bottleneck is not merely access to historical evidence, but converting that evidence into the state required by the continuation.

The category-wise results reveal three failure modes. In Past-visible cases, retrieved frames help, but the model must still decide whether a past observation remains current. In Occluded-process cases, the target state is produced by a hidden update and never appears as a reusable target frame. In Complex-transition cases, similarity to past observations is a weak cue for spatial transfer, quantity change, color mixing, or physical-state change. Figure~\ref{fig:qualitative} makes these failures concrete: some baselines fail to tilt the box toward the camera, so the state-checking action is incomplete; others open the box but show an empty interior despite the apple being placed inside in the previous segment; still others hallucinate an ambiguous or mismatched object. \stateagent avoids these failures by predicting that the apple should remain inside the box and grounding that target state as a future end frame before rendering.

\subsection{Ablation on State Conditioning}

Table~\ref{tab:ablation} shows that the gain is not from stronger hints alone. State-based prompt enhancement improves SES to 90.0, confirming that textual state hints help execute the requested action, but its SCS remains 61.9 because text lacks the visual grounding needed to bind the predicted state to entity appearance and so on. Direct end-frame editing provides such visual grounding and improves over several observation-memory baselines, but without the full state-update pipeline it still trails \stateagent. The full pipeline achieves the best SCS and SCS-All by coupling state prediction, future-frame grounding, and video rendering.

\subsection{State Consistency in Multi-Shot Story Generation}

\begin{figure}[t]
\centering
\includegraphics[width=0.96\linewidth]{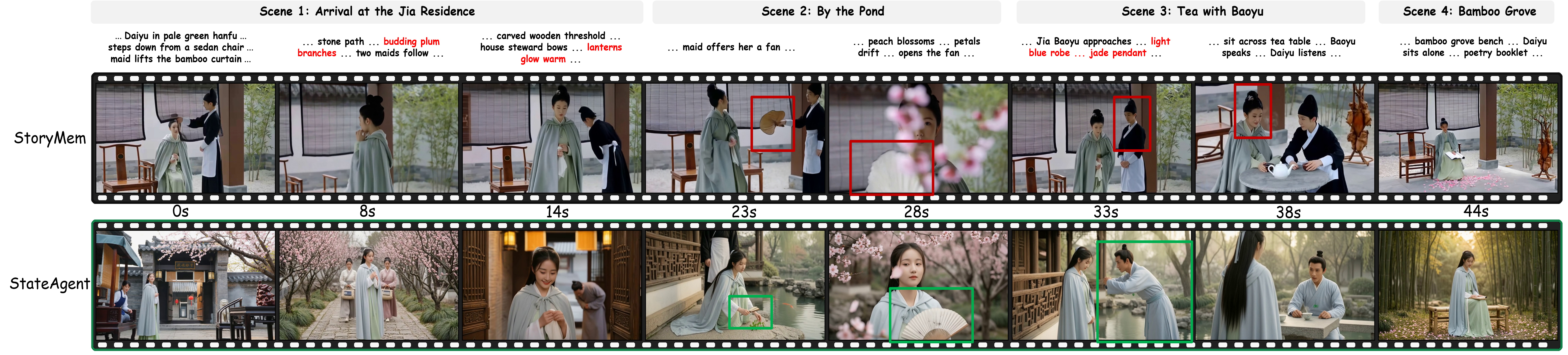}
\caption{Long story generation example of ST-Bench. The same story script is given to StoryMem and \stateagent.}
\label{fig:story_transfer}

\vspace{8pt}
\begin{minipage}[t]{0.49\linewidth}
\centering
\footnotesize
\setlength{\tabcolsep}{3pt}
\renewcommand{\arraystretch}{0.95}
\begin{tabular}{@{}lccc@{}}
\toprule
Variant & SES$\uparrow$ & SCS$\uparrow$ & SCS-All$\uparrow$ \\
\midrule
Base I2V & 78.0 & 41.6 & 32.5 \\
Direct end-frame Editing & 79.0 & 56.6 & 44.7 \\
State-based Prompt Enhancement & \underline{90.0} & \underline{61.9} & \underline{55.8} \\
Full \stateagent & \textbf{92.5} & \textbf{74.9} & \textbf{69.3} \\
\bottomrule
\end{tabular}
\captionof{table}{Ablation on state conditioning.}
\label{tab:ablation}
\end{minipage}\hfill
\begin{minipage}[t]{0.49\linewidth}
\centering
\footnotesize
\setlength{\tabcolsep}{4pt}
\begin{tabular}{@{}lccc@{}}
\toprule
Method & Aesthetic & PF & Consistency \\
\midrule
StoryMem & 5.81 & 0.222 & 0.586 \\
\stateagent & \textbf{6.35} & \textbf{0.230} & \textbf{0.594} \\
\bottomrule
\end{tabular}
\captionof{table}{ST-Bench results for multi-shot story generation.}
\label{tab:story_transfer}
\end{minipage}
\end{figure}

To test whether our method remains useful beyond the controlled unit setting of \statebench, we evaluate multi-shot story generation under the ST-Bench protocol from StoryMem~\citep{storymem}. ST-Bench consists of 1-minute long story scripts, shot-level video prompts, and scene-cut indicators. Following this protocol, each method generates the story shot by shot from the same script-level inputs. We evaluate aesthetic quality, prompt following, and cross-shot consistency following StoryMem.

Table~\ref{tab:story_transfer} reports the ST-Bench metrics, while Figure~\ref{fig:story_transfer} provides a qualitative comparison. \stateagent achieves improvements across all metrics. The qualitative example further shows that explicit state update helps preserve entity states and object appearances while following scene-level prompt changes.

The qualitative example reveals the limitation of a fixed visual memory bank. StoryMem stores only a limited number of memory frames, with early key frames acting as a memory sink and recent frames capturing short-term dynamics. This can make early characters and scenes repeatedly dominate later generations, causing repeated backgrounds, incorrect binding of new characters to earlier appearances, and loss of small but state-relevant objects. For example, a palm-leaf fan may drift into a paper fan when its appearance is diluted in memory. \stateagent instead maintains explicit entity states and updates them under each prompt, allowing it to follow scene changes, preserve object appearance and state, and avoid incorrect character binding.

\section{Related Work}

\paragraph{Video continuation and visual memory.}
Video generators achieve strong short-term quality with DiT backbones~\citep{vdm,imagenvideo,makeavideo,dit,latte,wan,hunyuanvideo}. Long-video systems extend them through chunkwise continuation~\citep{streamingt2v}, selected keyframes~\citep{storydiffusion,storymem}, entity or spatial memories~\citep{videomemory,worldmem,vmem,spatialmemory}, and KV caches~\citep{contextforcing,causalforcing,causalcine}. These interfaces preserve useful observations such as appearance, identity and so on. Our focus is complementary: converting historical observations into the current world state required by the next segment.

\paragraph{State evolution and video world models.}
Recent systems have started to explore evolving unseen states. PAN models long-horizon dynamics with LLM latent features and a video decoder~\citep{pan}, but its state lacks explicit appearance binding for continuation. LiveWorld maintains a persistent 4D spatiotemporal memory for out-of-sight dynamics~\citep{liveworld}, but this powerful scene-level representation is costly. ReMind trains video generators to use KV-cache mechanisms as dynamic memory for hidden-state evolution~\citep{remind}, showing that specialized training can improve this ability. Our work instead evaluates this capability in current continuation systems and tests an explicit state carrier without retraining the video generator.

\paragraph{Benchmarks for generation, memory, and state reasoning.}
Video-generation benchmarks initially focused on visual quality, text alignment and so on~\citep{vbench,evalcrafter,t2vcompbench,videophy,phygenbench,worldbench,NarrLV,Omni-worldbench}. Recent benchmarks have begun to shift toward memory, state reasoning, and out-of-sight state evolution. WorldReasonBench evaluates whether video generators can predict future world states from an initial condition and action~\citep{worldreasonbench}. MIND studies memory consistency and action control in video world models~\citep{mind}. STEVO-Bench studies whether natural state evolution can continue when observation is interrupted by occluders, lights-off intervals, or look-away trajectories, but state setup, hiding, evolution, and reveal are coupled within one generation, making failures hard to attribute~\citep{stevobench}. MemoBench studies revisiting behavior in dynamically changing environments, but its focus is camera-controlled video generation~\citep{memobench}. Despite this progress, world-state handoff in cross-segment video continuation remains underexplored: a historical video establishes a world state, and a later prompt asks the next generated segment to use or change it.

\section{Conclusion}

We introduced cross-segment state handoff as a requirement distinct from observation memory in video continuation. We proposed \statebench to test whether continuations preserve and update states established by previous segments across past-visible, occluded-process, and complex-transition cases. We also presented \stateagent, which explicitly predicts the continuation state, grounds it as a future end frame, and renders the next segment. Experiments show that stronger visual memory alone does not make current generators reliably infer world state; coherent continuation benefits from an explicit state belief that can be updated, grounded to appearance, and carried across generation boundaries, with supporting evidence from ST-Bench story generation.

\bibliographystyle{plainnat}
\bibliography{references}

\appendix
\thispagestyle{fancy}
\section{Experimental Details}

This section provides more details: task metadata, implementation details of \statebench and \stateagent, settings of baselines.

\subsection{StateBench Data Construction}

We construct \statebench by first defining fine-grained state types and then generating paired prompts for the previous segment and the continuation segment. The previous prompt creates a state-establishing event, while the continuation prompt asks for an action that exposes, uses, or transforms the resulting state without naming the expected answer. Previous videos are generated with Wan2.2-T2V/I2V-plus at 480p resolution. For tasks requiring multi-step setup, intermediate setup segments are generated sequentially by conditioning on the previous segment boundary.

All generated histories are manually filtered before evaluation. A retained case must satisfy four conditions: (1) the state-changing event is completed, (2) the boundary frame does not directly reveal the task-relevant state, (3) the continuation prompt does not leak the target answer, and (4) the case can be evaluated by concrete exposure/action and state-consistency checklists. Cases that fail these checks are regenerated or removed. The final benchmark contains 85 past-visible, 65 occluded-process, and 50 complex-transition cases.

\subsection{Case Schema and State Types}

Each \statebench item is stored as a structured record containing task metadata, prompts, state annotations, and evaluation checklists:
\begin{quote}
\small
\texttt{id}: unique task identifier.\\
\texttt{category}: one of past-visible, occluded-process, or complex-transition.\\
\texttt{state\_type}: coarse state family, such as existence, quantity/degree, spatial, physical state, surface/appearance, or agent state.\\
\texttt{fine\_type}: fine-grained state type.\\
\texttt{previous\_video}: previous segment containing the state-establishing event.\\
\texttt{previous\_prompt}: source prompt used to generate the previous segment.\\
\texttt{state\_after\_previous}: annotated state at the historical end frame.\\
\texttt{continuation\_prompt}: prompt for the next segment.\\
\texttt{target\_state}: state that should be reflected in the continuation.\\
\texttt{evaluation}: exposure/action checklist and state-consistency checklist.
\end{quote}

Table~\ref{tab:appendix_state_types} summarizes the fine-grained state types. These types are controlled probes rather than a complete ontology of world state.

\begin{table}[t]
\centering
\small
\setlength{\tabcolsep}{4pt}
\renewcommand{\arraystretch}{0.95}
\begin{tabular}{@{}p{0.17\textwidth}p{0.31\textwidth}p{0.44\textwidth}@{}}
\toprule
State family & Fine-grained types & Examples of state reasoning \\
\midrule
Existence & containment, occlusion, coverage, exit\_frame, departure & Object remains inside a container; object is hidden, covered, or leaves the scene. \\
Quantity/Degree & consumption, addition, division, emptying & Amount increases, decreases, is consumed, divided, or emptied. \\
Spatial & displacement, swap, flip & Object moves, swaps position, flips, or is tracked under occlusion. \\
Physical state & open\_close, fold, breakage, melting, cooking & Object opens or closes, folds, breaks, melts, or changes through heating/cooking. \\
Surface/Appearance & writing, staining, labeling, wearing & Text, stains, labels, or worn objects change visible appearance. \\
Agent state & held\_object, posture, costume\_change & Agent holds a different object, changes posture, or changes visible outfit or appearance. \\
\bottomrule
\end{tabular}
\caption{Fine-grained state types used in \statebench.}
\label{tab:appendix_state_types}
\end{table}

\begin{table}[t]
\centering
\small
\setlength{\tabcolsep}{4pt}
\renewcommand{\arraystretch}{0.95}
\begin{tabular}{@{}p{0.24\textwidth}p{0.25\textwidth}p{0.43\textwidth}@{}}
\toprule
Component / Baseline & Model / Interface & Notes \\
\midrule
Previous-video generation & Wan2.2-T2V/I2V-plus & 480p, 16 FPS, manually filtered \\
Final-frame I2V baseline & Wan2.2-I2V-A14B & Conditions on the historical end frame \\
Recent-frame baseline & LongCat-Video & Video-prefix continuation interface \\
Memory baseline & StoryMem~\citep{storymem} & Official memory-based continuation interface \\
Full key-frame baseline & StoryMem + full KFs & Uniform historical KFs: 6 frames per 5 seconds; max 10 frames in the StoryMem memory bank \\
Autoregressive baseline & MAGI-1~\citep{magi1} & Native video continuation baseline \\
VLM components and judge & Qwen3.5-Plus & State parsing, update, prompting, and evaluation \\
Future-frame grounding & Wan2.7-Image & Image generation/editing API \\
Keyframe-to-video rendering & Wan2.2-KF2V-Flash & Historical end frame + future end frame \\
\bottomrule
\end{tabular}
\caption{Base models and interfaces used in our experiments.}
\label{tab:base_models}
\end{table}

\begin{table}[t]
\centering
\small
\setlength{\tabcolsep}{4pt}
\renewcommand{\arraystretch}{0.95}
\begin{tabular}{@{}lrrrrrrrr@{}}
\toprule
Dimension & Judgments & Score & Ours win & Baseline win & Tie & Win w/o ties & Maj. ours & Maj. agree \\
\midrule
Aesthetic quality & 450 & 74.7 & 61.3 & 12.0 & 26.7 & 83.6 & 68.8 & 74.0 \\
State reveal success & 450 & 81.0 & 71.8 & 9.8 & 18.4 & 88.0 & 80.0 & 81.0 \\
State consistency & 450 & 85.6 & 81.3 & 10.2 & 8.4 & 88.8 & 82.5 & 90.0 \\
\midrule
All & 1,350 & 80.4 & 71.5 & 10.7 & 17.9 & 87.0 & 77.1 & 81.7 \\
\bottomrule
\end{tabular}
\caption{Pairwise human evaluation by dimension. Score uses 1/0.5/0 for ours win/tie/baseline win. Win rates, tie rates, majority preference, and majority agreement are reported in percentages.}
\label{tab:user_study_dimension}
\end{table}

\begin{table}[t]
\centering
\small
\setlength{\tabcolsep}{3pt}
\renewcommand{\arraystretch}{0.95}
\begin{tabular}{@{}lrrrrr@{}}
\toprule
Baseline & N & Score & Ours win & Base win & Tie \\
\midrule
Final-frame I2V & 375 & 77.5 & 66.1 & 11.2 & 22.7 \\
LongCat-Video & 318 & 80.0 & 71.4 & 11.3 & 17.3 \\
StoryMem & 369 & 72.1 & 61.0 & 16.8 & 22.2 \\
MAGI-1 & 288 & 95.3 & 92.0 & 1.4 & 6.6 \\
\bottomrule
\end{tabular}
\caption{Pairwise human evaluation by baseline. Scores and rates are aggregated over all three dimensions and reported in percentages.}
\label{tab:user_study_baseline}
\end{table}

\subsection{Checklist Evaluation and VLM Judge}

In this section, we describe the checklist instantiation. Each task contains three question groups: \emph{reveal achieved}, \emph{state correct}, and \emph{no violation}. The first group contains one action-completion question. The second group checks whether the exposed or used state matches the annotated target state. The third group checks that the reveal does not introduce unsupported objects, duplicated targets, identity changes, contradictory attributes, or impossible contents.

Some checklist questions require appearance or identity comparison against the history. For these cases, we provide the judge with task-specific reference frames in which the relevant entity appearance can be clearly bound to the historical evidence. These references help the evaluator decide whether the revealed entity in the continuation matches the entity established in the previous video.

All VLM-based evaluation uses Qwen3.5-Plus. The judge receives the previous video, prompts, generated continuation, task-specific questions, and any referenced historical frames. We use a compact JSON-format instruction:
\begin{promptbox}[VLM Judge]
You are a video state evaluator. Watch the generated continuation together with the historical video and any referenced historical frames. Answer each checklist question with yes or no based only on visible evidence and the provided historical context. For appearance-sensitive questions, compare the revealed entity with the referenced historical frames. Return a JSON object with an answer and a brief reason for each question. Do not add extra text outside JSON.
\end{promptbox}

\subsection{StateAgent Implementation Details}

All VLM-based components in \stateagent use Qwen3.5-Plus. The image generation/editing module uses Wan2.7-Image, and the video renderer uses Wan2.2-KF2V-Flash. The source prompt $p_t$ is used to bind textual names to visual entities; state parsing is grounded in observed video evidence.

\paragraph{State graph fields.}
Entity nodes store identity and type, together with open-ended appearance description, visibility, location, physical condition, and concise state description inferred by the VLM from the observed video. Relation edges are also open-ended and may describe containment, location, attachment, surface/material state, quantity, or interaction. Each entity is assigned a clear reference frame from the previous video when available, so that future-frame grounding can preserve appearance.

\paragraph{Core prompts.}
We use structured prompts for the VLM modules. Below we report only the core logic of the prompts. Engineering-oriented instructions, such as exact output schemas, formatting constraints, retry rules, and implementation-specific checks, are omitted for readability and will be included in the released code.

\begin{promptbox}[Entity Binding]
You are an entity extractor. Extract only entity names, entity types, and explicitly mentioned appearance attributes from the source prompt. Do not infer hidden state, future state, or task answers. Use the extracted entities only to bind textual names to visual entities observed in the video.
\end{promptbox}

\begin{promptbox}[State Parsing]
You are a video state observer. Analyze the observed video evidence and return the actual state of each entity at the current boundary. Report visible entities, hidden entities, locations, container open or closed states, physical conditions, and relations such as inside, holding, wearing, covered by, behind, or occluded by. The source prompt may help bind names to visual entities, but state parsing must follow what is observed in the video. If the prompt and video disagree, report the observed video state.
\end{promptbox}

\begin{promptbox}[State Update]
You are a video state predictor. Given the current world state and the next-shot prompt, predict the complete world state at the end of the next shot. Treat the prompt as an action condition. Update each entity's visibility, location, physical condition, appearance description, concise state description, and open-ended relations such as containment, attachment, surface/material state, quantity, and agent-object interaction. Carefully distinguish physical existence from camera visibility: an entity can exist while remaining hidden from the current viewpoint.
\end{promptbox}

\begin{promptbox}[Editing Prompt Writer]
You are an expert at writing image-editing prompts. The last numbered image is the current frame to edit, and earlier numbered images are appearance references. Describe the expected result image in natural photographic language. State what should be visible after the action, refer to earlier images when matching entity appearance is needed, and preserve background, lighting, camera, and unchanged objects when the viewpoint does not change. If the prompt changes viewpoint, describe the full scene from the new viewpoint.
\end{promptbox}

\paragraph{Reference selection.}
For future-frame grounding, \stateagent includes references only for entities involved in the predicted update or entities that are hidden at the boundary but should become visible in the continuation.

\subsection{Base Models and Generation Settings}

Table~\ref{tab:base_models} lists the model interfaces used in our experiments. All methods are evaluated under matched continuation prompts and comparable output settings within each model family. Default API or official implementation settings are used unless otherwise specified.

The StoryMem + full key-frame baseline uniformly samples key frames from the historical video, using six frames per five seconds while respecting the StoryMem memory bank limit of 10 input frames. It tests whether dense visual evidence alone is sufficient for state update under the current continuation interface.

\section{User Study}

We also conduct a double-blind pairwise human evaluation. The study focuses on the generated continuation rather than the historical context. Each trial compares \stateagent with one baseline; the display position is randomized and method names are hidden from participants. The preceding historical video is shown as reference, and participants are instructed to judge only the final continuation segment.

We compare against four baselines: final-frame I2V, LongCat-Video, StoryMem, and MAGI-1. Each participant is randomly assigned 30 comparisons. We collect 15 complete submissions, resulting in 450 pairwise answers and 1,350 dimension-level judgments. For each comparison, participants choose which result is better, or whether the two are tied, along three dimensions: aesthetic quality, state reveal success, and state consistency.

We decode each preferred-side/tie response using the hidden display order. For each dimension-level judgment, \stateagent receives score 1 if it is preferred, 0.5 for a tie, and 0 if the baseline is preferred. We report the resulting score rate, raw win/tie rates, and win rate excluding ties. We also aggregate votes by scenario, baseline, and dimension to compute majority preference and majority agreement. Majority agreement is the fraction of votes assigned to the most selected option in each aggregated cell.

Table~\ref{tab:user_study_dimension} shows that humans prefer \stateagent most strongly on the state-related dimensions. The gain is largest for state consistency, where \stateagent obtains an 85.6 score rate, an 81.3 raw win rate, and an 88.8 win rate after excluding ties. Majority agreement is also highest on state consistency, suggesting that participants form clearer consensus when judging whether the revealed state matches the historical event. Aesthetic quality shows a smaller margin and more ties, which is expected because it is more subjective and because several methods share similar visual generation priors.

Table~\ref{tab:user_study_baseline} breaks down the results by baseline. The advantage is consistent across all comparisons, with the smallest margin against StoryMem and the largest margin against MAGI-1. This pattern matches the automatic evaluation: memory-based baselines can preserve some visual evidence, but participants still prefer \stateagent when the continuation must reveal or use the state established by the history.

\section{Additional Results}

\subsection{Additional StateBench Qualitative Results}

\begin{figure}[p]
\centering
\includegraphics[width=0.9\linewidth]{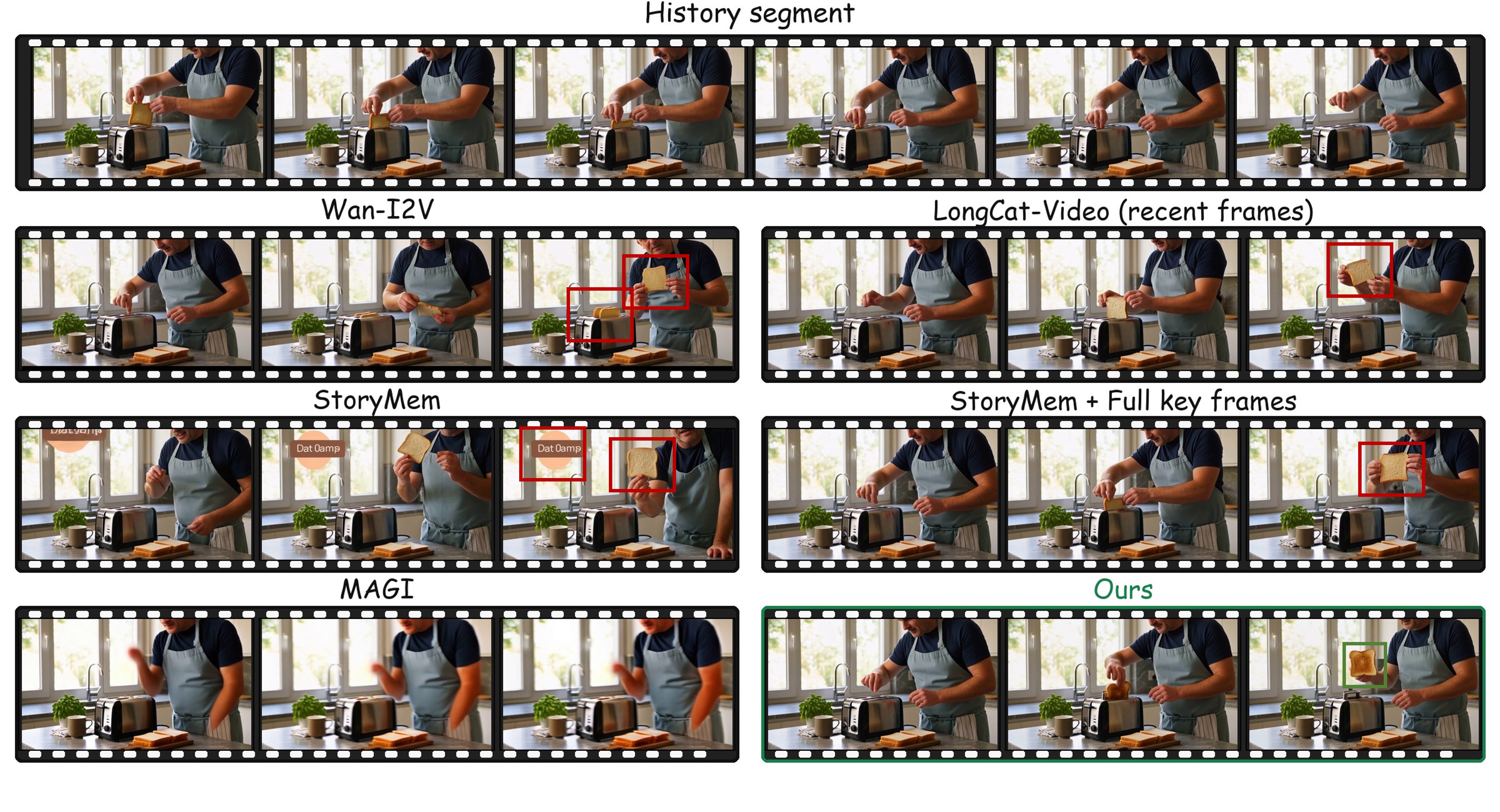}
\caption{\textbf{Additional qualitative comparison.} \emph{History prompt:} A single continuous realistic video on a kitchen counter. A man stands behind the counter with a slice of white bread and a toaster, all clearly visible. He places the white bread vertically into the toaster slot, then presses the toaster lever down; the bread fully enters the machine. Fixed camera, medium shot, bright indoor lighting. \emph{Continuation prompt:} A single continuous realistic video on a kitchen counter, same toaster. Time is up, the toaster pops. The man takes the bread slice out and holds it up to the camera. Fixed camera, medium shot, bright indoor lighting.}
\label{fig:appendix_result_a1}
\end{figure}

\begin{figure}[p]
\centering
\includegraphics[width=0.9\linewidth]{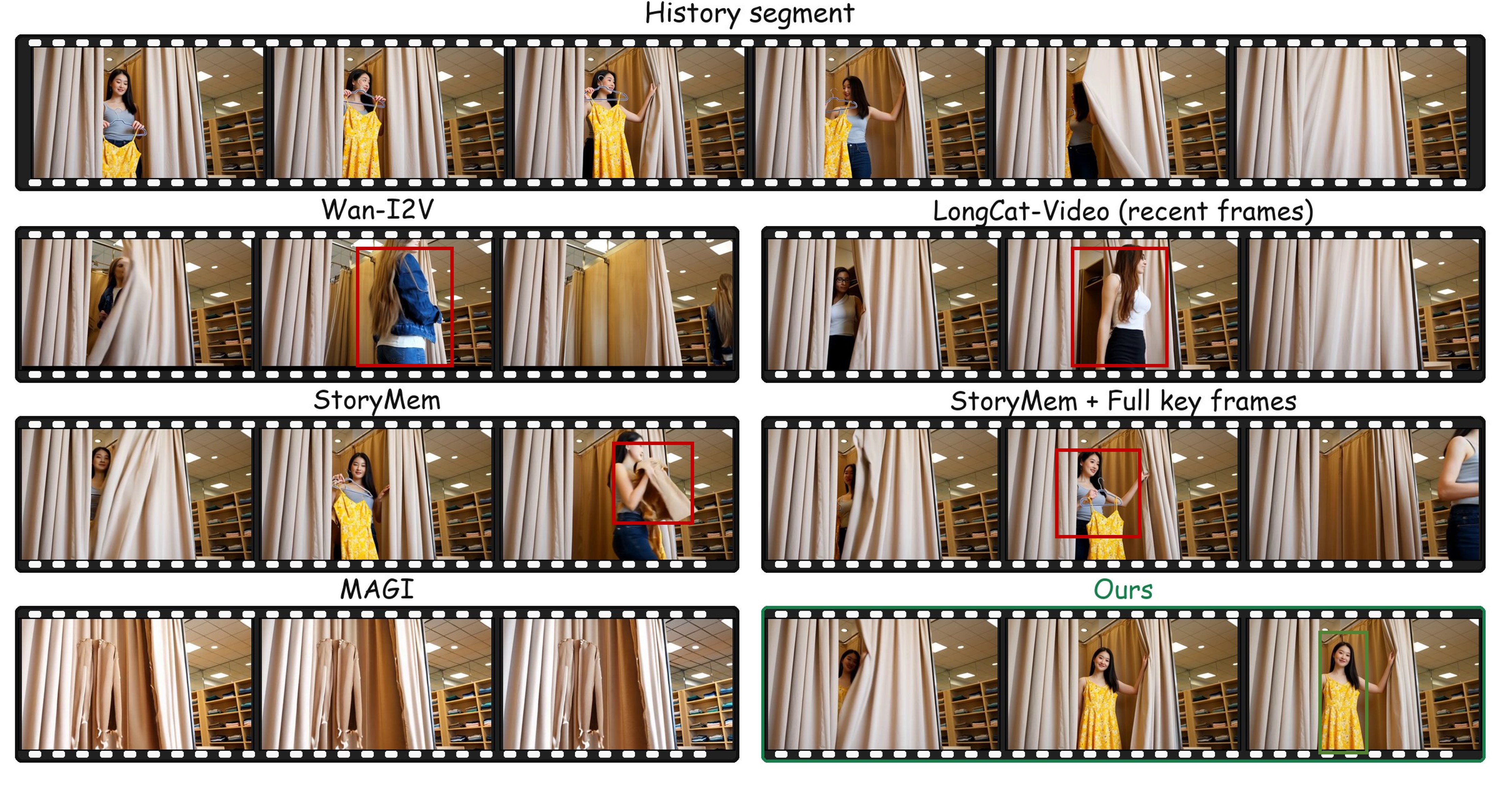}
\caption{\textbf{Additional qualitative comparison.} \emph{History prompt:} A single continuous realistic video. The camera faces a fitting room with a curtain. A woman in a plain gray tank top holds up a hanger with a yellow floral dress to show it to the camera, then carries it into the fitting room and pulls the curtain closed. Fixed camera, medium shot, bright indoor lighting. \emph{Continuation prompt:} A single continuous realistic video. The curtain of the fitting room is pulled open. She tried on her clothes and walked out. Fixed camera, medium shot, bright indoor lighting.}
\label{fig:appendix_result_d2}
\end{figure}

\begin{figure}[p]
\centering
\includegraphics[width=0.9\linewidth]{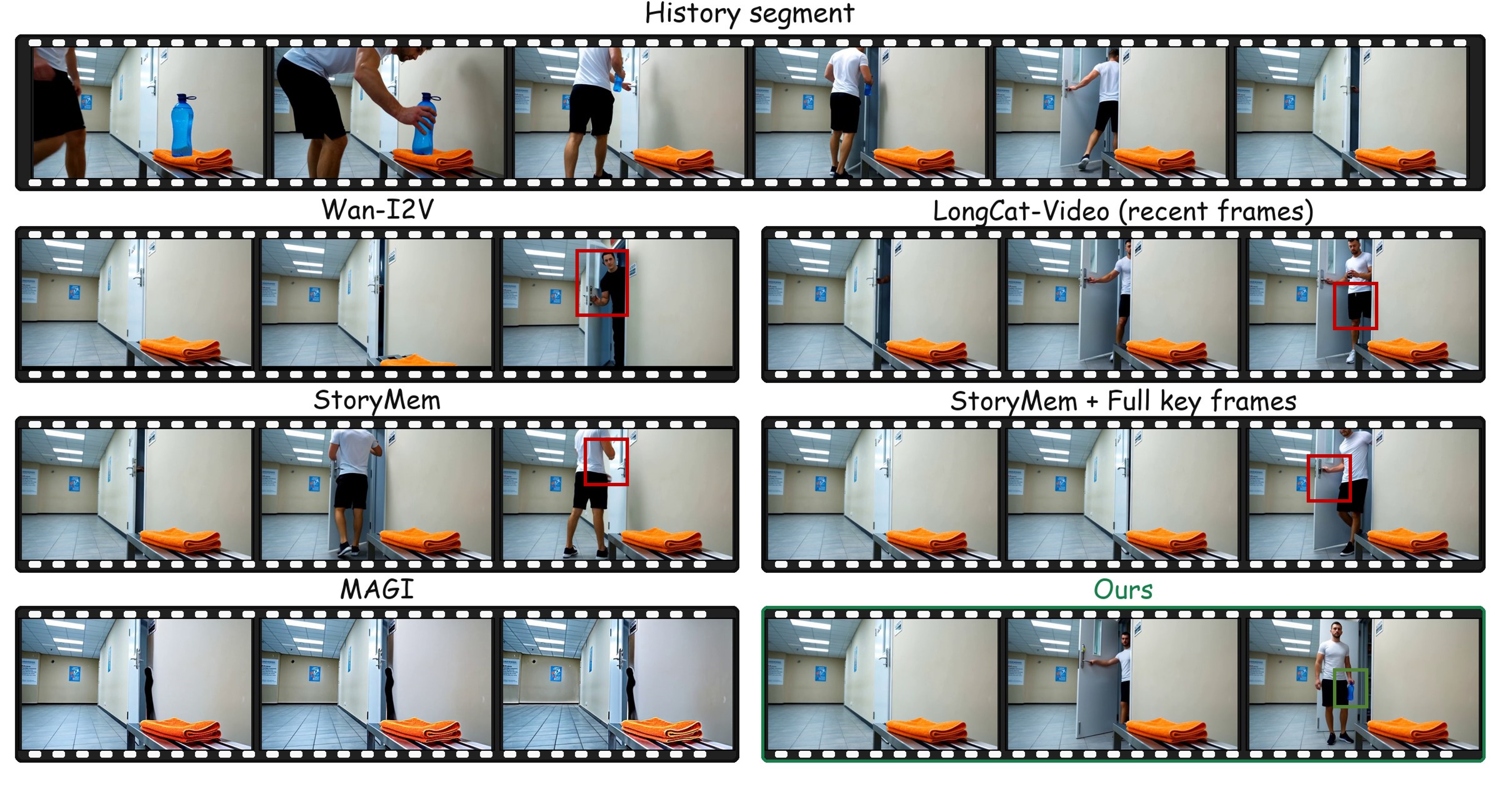}
\caption{\textbf{Additional qualitative comparison.} \emph{History prompt:} A single continuous realistic video in a gym hallway with a closed door. A blue water bottle and an orange towel rest on a bench. A man picks up the blue water bottle, leaving the towel, walks through the doorway into the locker room, and closes the door behind him. Fixed camera, medium shot, bright indoor lighting. \emph{Continuation prompt:} A single continuous realistic video in a gym hallway. The door opens, revealing the man standing inside. Fixed camera, medium shot, bright indoor lighting.}
\label{fig:appendix_result_h5}
\end{figure}

\begin{figure}[p]
\centering
\includegraphics[width=0.9\linewidth]{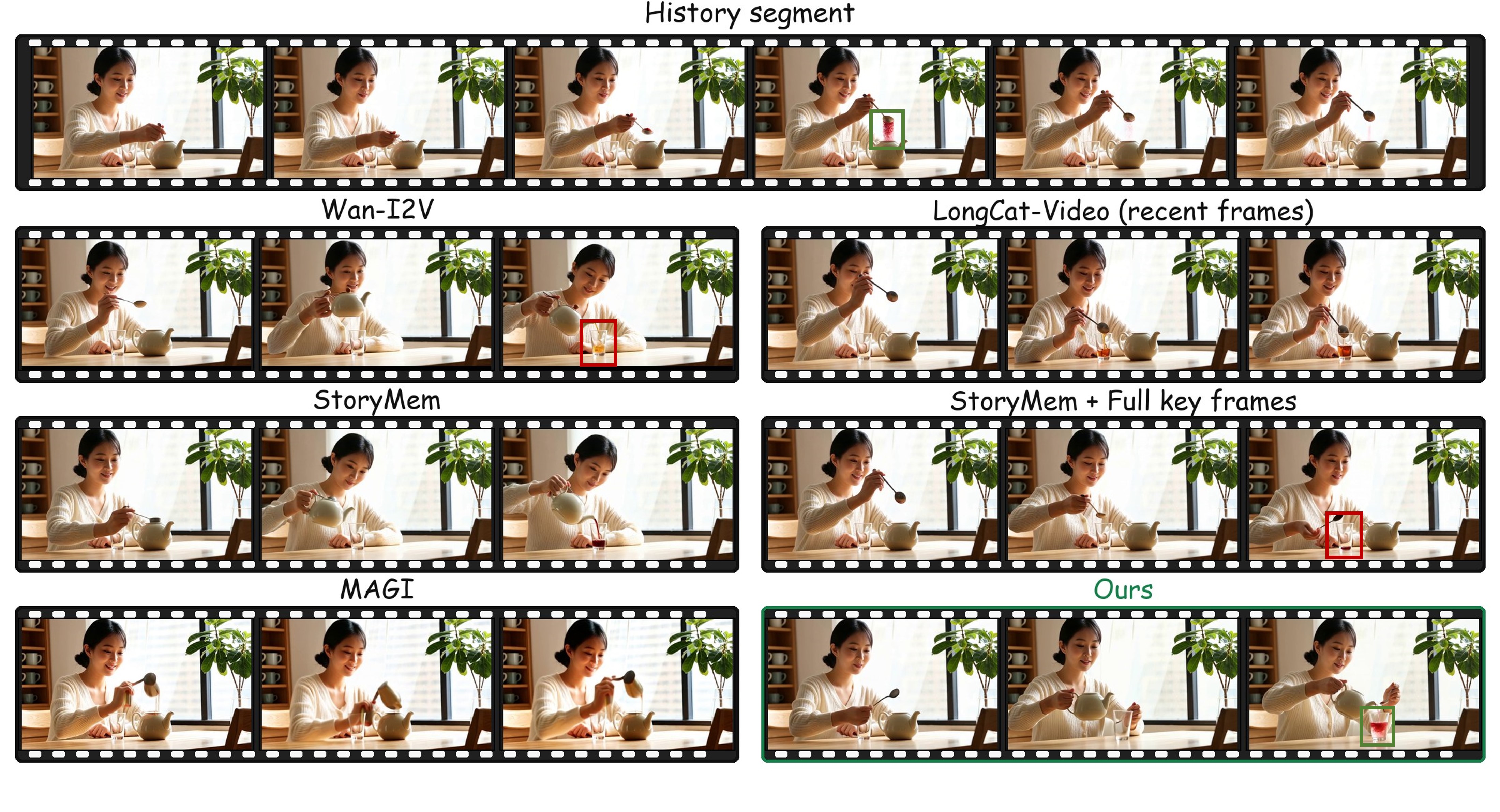}
\caption{\textbf{Additional qualitative comparison.} \emph{History prompt:} A single continuous realistic video on a table. A woman sits at the table with an empty transparent glass cup and an opaque teapot already filled with clear water, both clearly visible. She scoops a spoonful of red fruit-punch drink powder and drops it into the teapot; the liquid inside the opaque teapot stays hidden from the camera. Fixed camera, medium shot, bright indoor lighting. \emph{Continuation prompt:} A single continuous realistic video on a table. The woman lifts the teapot and pours its liquid into the transparent glass cup. Fixed camera, medium shot, bright indoor lighting.}
\label{fig:appendix_result_l7}
\end{figure}

\begin{figure}[p]
\centering
\includegraphics[width=0.9\linewidth]{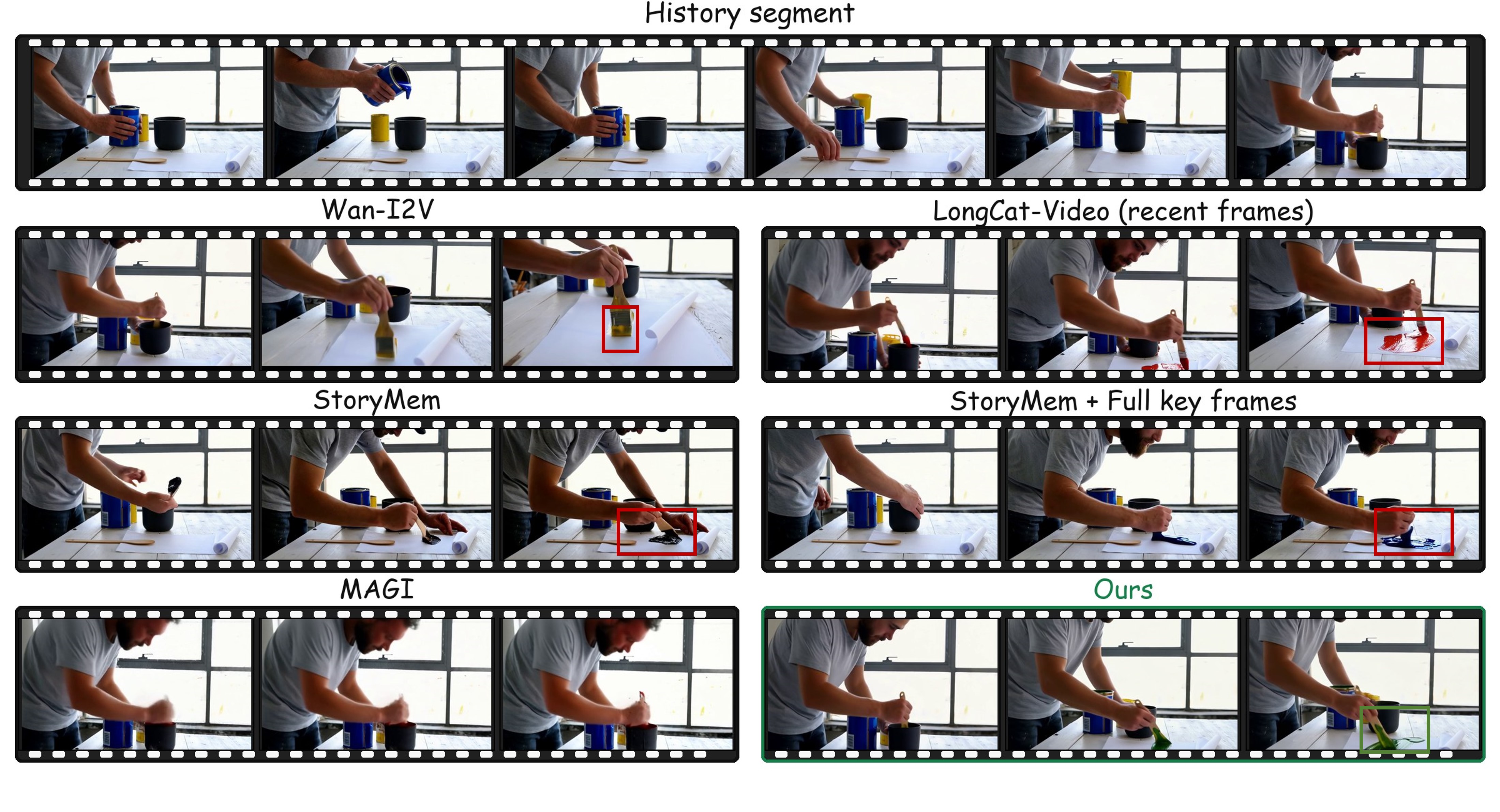}
\caption{\textbf{Additional qualitative comparison.} \emph{History prompt:} A single continuous realistic video on a table. A man stands at the table with a can of blue paint, a can of yellow paint, a sheet of white paper, a wooden stirring stick, and a completely opaque black mixing cup. He pours blue paint into the opaque cup, then pours yellow paint into the same cup, and stirs thoroughly with the stick until mixing is complete; the cup remains fully opaque. Fixed camera, medium shot, bright indoor lighting. \emph{Continuation prompt:} A single continuous realistic video on the same table. The man pulls the stirring stick out of the opaque cup and smears the paint from the stick onto the white paper, revealing the mixed paint color on the paper. Fixed camera, medium shot, bright indoor lighting.}
\label{fig:appendix_result_m1}
\end{figure}

\begin{figure}[p]
\centering
\includegraphics[width=0.9\linewidth]{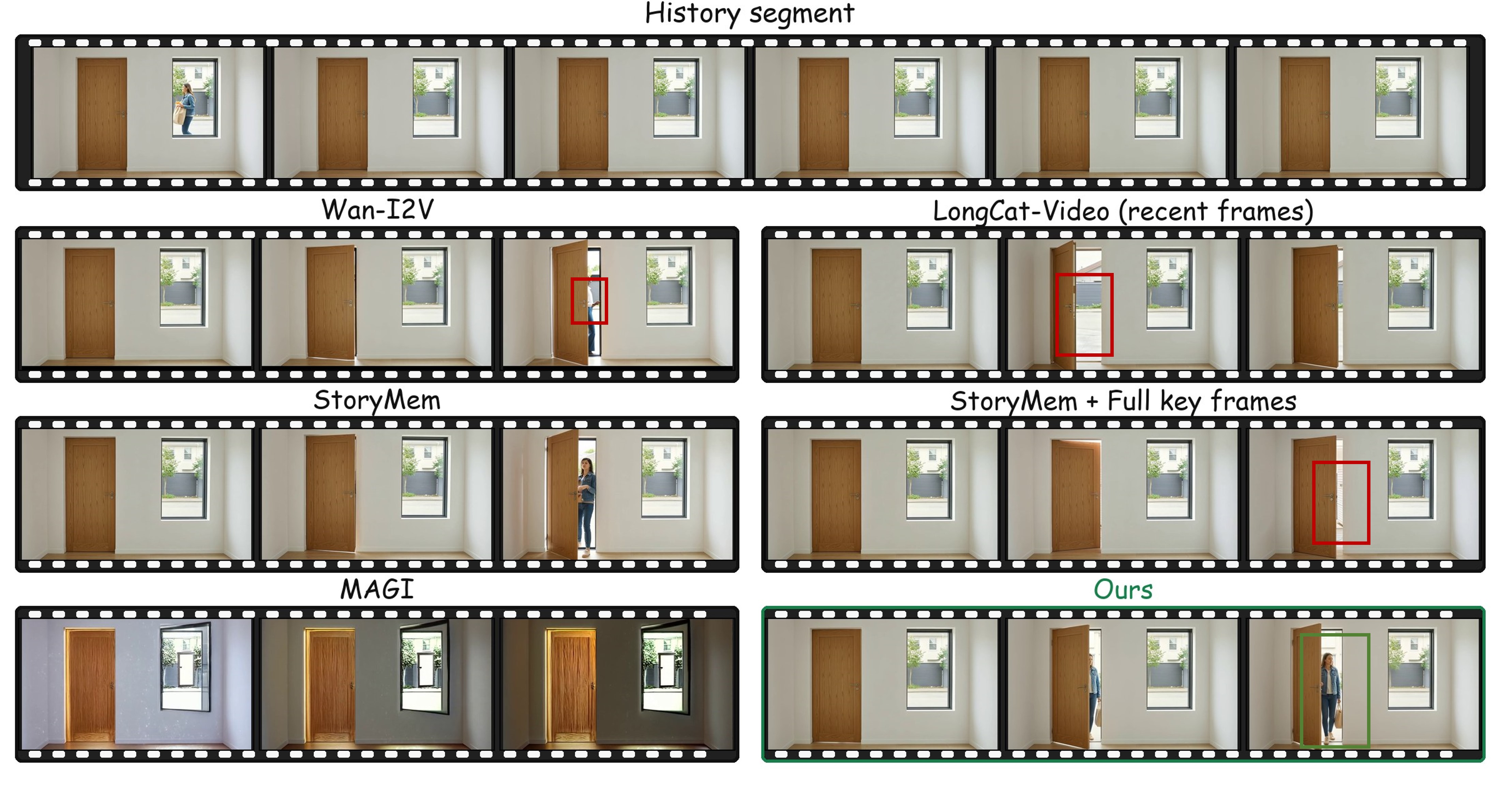}
\caption{\textbf{Additional qualitative comparison.} \emph{History prompt:} A single continuous realistic video, fixed camera inside a room facing a closed front door with a window beside it. Through the window, a woman in a denim jacket holding a grocery bag is seen walking up from outside toward the door. She steps up to the door and stops, now hidden behind the closed door and no longer visible through the window. Fixed camera, medium shot, bright indoor lighting. \emph{Continuation prompt:} A single continuous realistic video, fixed camera inside the room. The front door opens. Fixed camera, medium shot, bright indoor lighting.}
\label{fig:appendix_result_o10}
\end{figure}

\begin{figure}[p]
\centering
\includegraphics[width=0.9\linewidth]{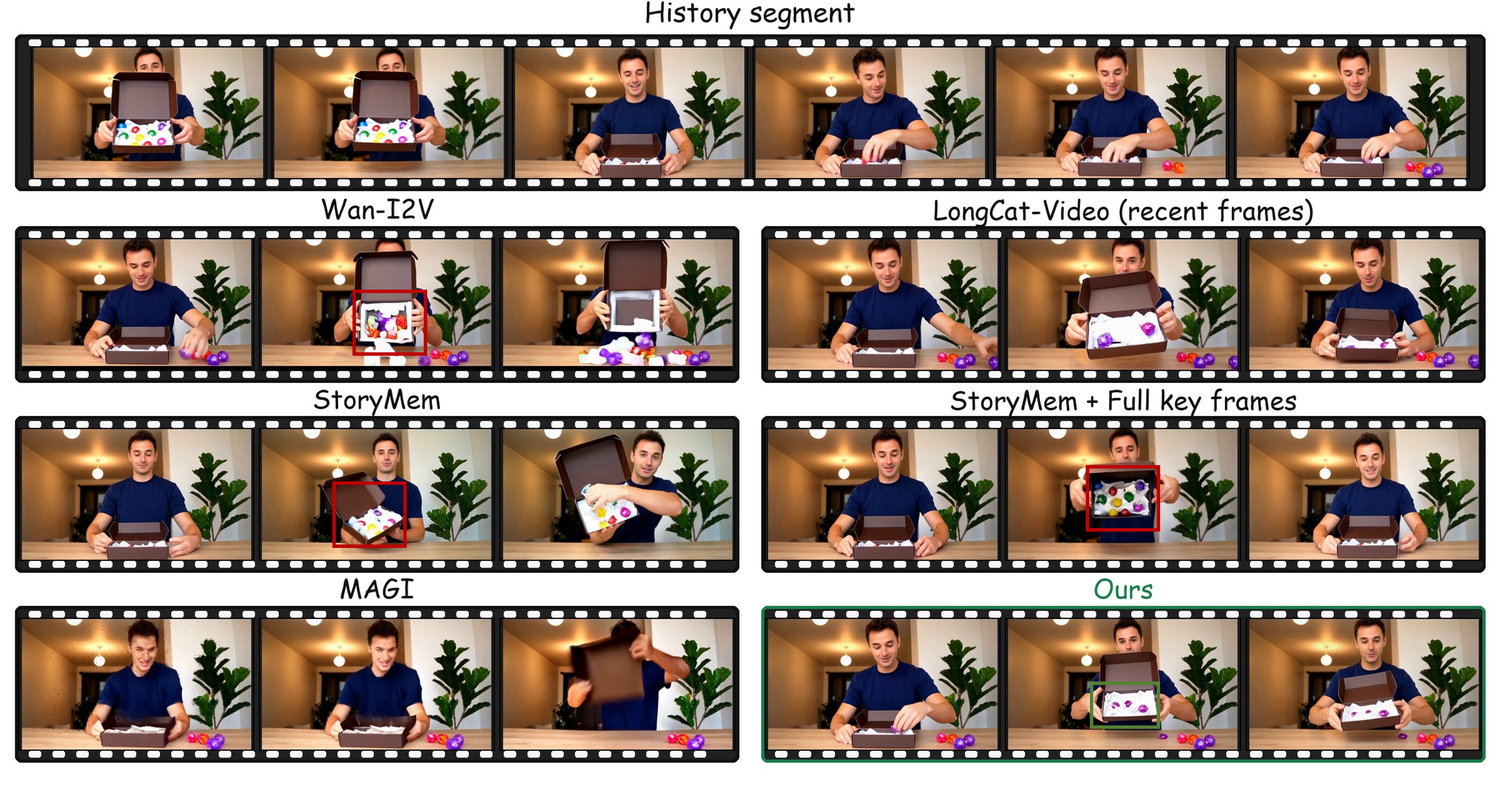}
\caption{\textbf{Additional qualitative comparison.} \emph{History prompt:} A single continuous realistic video on a table. A man stands behind the table with an open-top opaque cardboard box with tall side walls. He tilts the box toward the camera to show nine wrapped candies inside, then places the box upright so the camera cannot see inside. He then reaches in from above, takes out five candies, and places them on the table beside the box; the camera never sees the remaining candies. Fixed camera, medium shot, bright indoor lighting. \emph{Continuation prompt:} A single continuous realistic video on the same table. The man lifts the open-top cardboard box and tilts it toward the camera to show exactly what remains inside the box. Fixed camera, medium shot, bright indoor lighting.}
\label{fig:appendix_result_q6}
\end{figure}

\begin{figure}[p]
\centering
\includegraphics[width=0.9\linewidth]{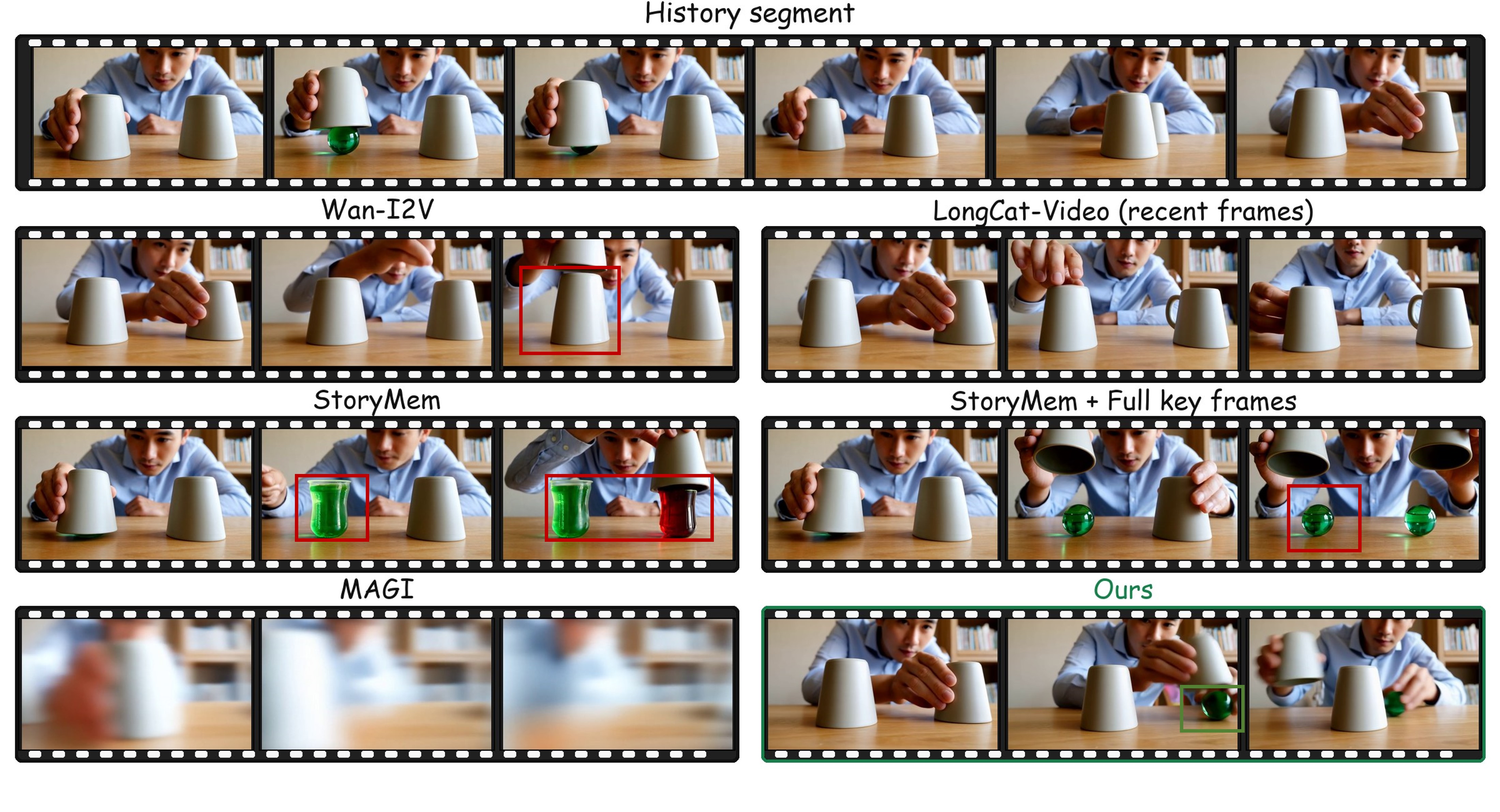}
\caption{\textbf{Additional qualitative comparison.} \emph{History prompt:} A single continuous realistic video. On a wooden table, two opaque cups are placed upside down side by side. A man lifts the left cup, revealing a green glass marble underneath, then places it back. Then he slides the left cup in a semicircular path around the right cup to the far right; the cup slides on the table without leaving it, the right cup stays still. Fixed camera, medium shot, bright indoor lighting. \emph{Continuation prompt:} A single continuous realistic video, same wooden table. The man lifts both the left and right cups simultaneously, showing what is under each. Fixed camera, medium shot, bright indoor lighting.}
\label{fig:appendix_result_s3}
\end{figure}

Figures~\ref{fig:appendix_result_a1}--\ref{fig:appendix_result_s3} provide additional qualitative comparisons on \statebench. Each result uses the same layout as Figure~\ref{fig:qualitative}: the top row shows six uniformly sampled frames from the historical segment, and each method row shows three frames sampled from the generated final five seconds.

\section{Limitation}

One limitation is that \stateagent currently renders the continuation from the historical end frame to a predicted future end frame. When these two frames differ substantially, the generated video may contain shot transitions. Richer temporal constraints or intermediate state anchors may help address this limitation.


\end{document}